\documentclass[11pt,letterpaper]{article}
\usepackage{naaclhlt2013}
\usepackage{times}
\usepackage{latexsym}
\usepackage{graphicx}
\usepackage{amsfonts}
\usepackage{url,amsmath}
\usepackage[utf8]{inputenc}
\usepackage[czech,english]{babel}
\title{Exploiting Similarities among Languages for Machine Translation}

\author{
Tomas Mikolov\\
Google Inc.\\
Mountain View \\
\texttt{tmikolov@google.com} \\
\And
Quoc V. Le \\
Google Inc. \\
Mountain View \\
\texttt{qvl@google.com} \\
\And
Ilya Sutskever \\
Google Inc. \\
Mountain View \\
\texttt{ilyasu@google.com} \\
}

\begin{document}
\maketitle
\begin{abstract}
  Dictionaries and phrase tables are the basis of modern statistical
  machine translation systems. This paper develops a method that can
  automate the process of generating and extending dictionaries and
  phrase tables.  Our method can translate missing word and phrase
  entries by learning language structures based on large monolingual
  data and mapping between languages from small bilingual data. It
  uses distributed representation of words and learns a linear mapping
  between vector spaces of languages. Despite its simplicity, our
  method is surprisingly effective: we can achieve almost 90\%
  precision@5 for translation of words between English and
  Spanish. This method makes little assumption about the languages, so
  it can be used to extend and refine dictionaries and translation
  tables for any language pairs.

\end{abstract}

\section{Introduction}
Statistical machine translation systems have been developed for years
and became very successful in practice. These systems rely on
dictionaries and phrase tables which require much efforts to generate
and their performance is still far behind the performance of human
expert translators.  In this paper, we propose a technique for machine
translation that can automate the process of generating dictionaries
and phrase tables. Our method is based on distributed representations
and it has the potential to be complementary to the mainstream
techniques that rely mainly on the raw counts.

Our study found that it is possible to infer missing dictionary
entries using distributed representations of words and phrases. We
achieve this by learning a linear projection between vector spaces
that represent each language. The method consists of two simple
steps. First, we build monolingual models of languages using large
amounts of text.  Next, we use a small bilingual dictionary to learn a
linear projection between the languages. At the test time, we can
translate any word that has been seen in the monolingual corpora by
projecting its vector representation from the source language space to
the target language space. Once we obtain the vector in the target
language space, we output the most similar word vector as the
translation.

The representations of languages are learned using the distributed
Skip-gram or Continuous Bag-of-Words (CBOW) models recently proposed
by~\cite{mikolov}. These models learn word representations using a
simple neural network architecture that aims to predict the neighbors
of a word. Because of its simplicity, the Skip-gram and CBOW models
can be trained on a large amount of text data: our parallelized
implementation can learn a model from billions of words in hours.\footnote{The code
for training these models is available at \\ \url{https://code.google.com/p/word2vec/}}

\begin{figure*}[tbh]
\centering
\includegraphics[width=.9\columnwidth]{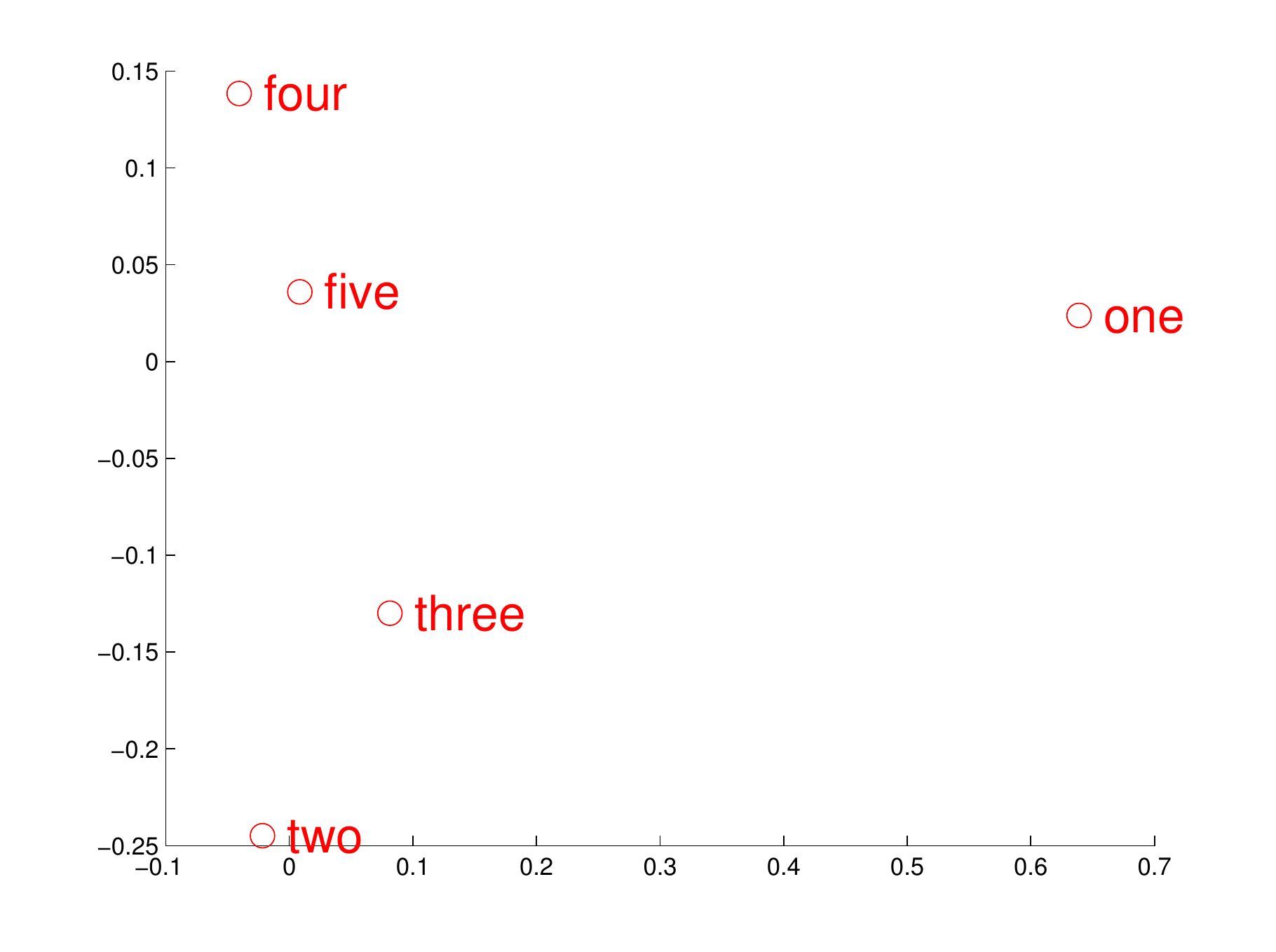}\Huge$\to$
\includegraphics[width=.9\columnwidth]{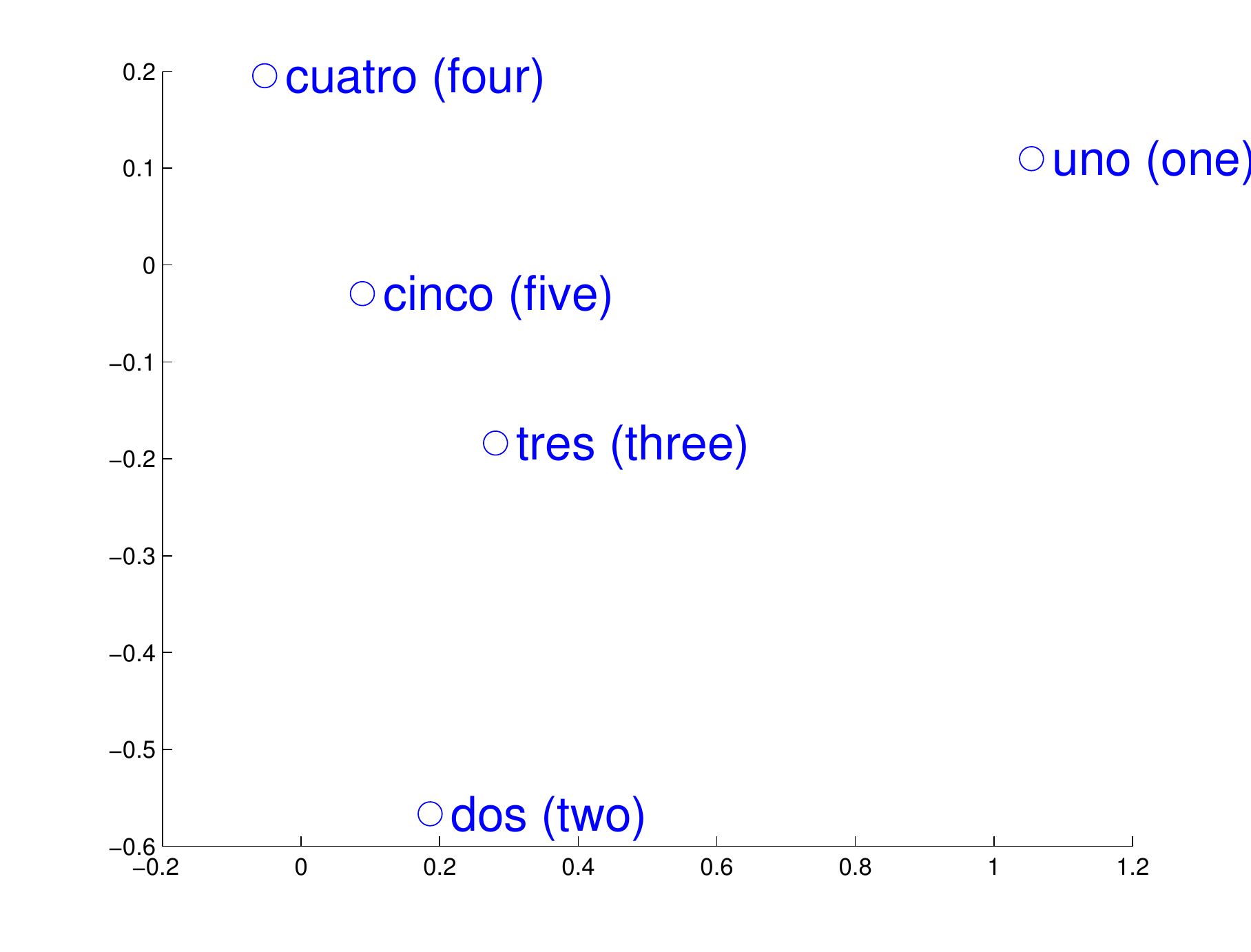}
\includegraphics[width=.9\columnwidth]{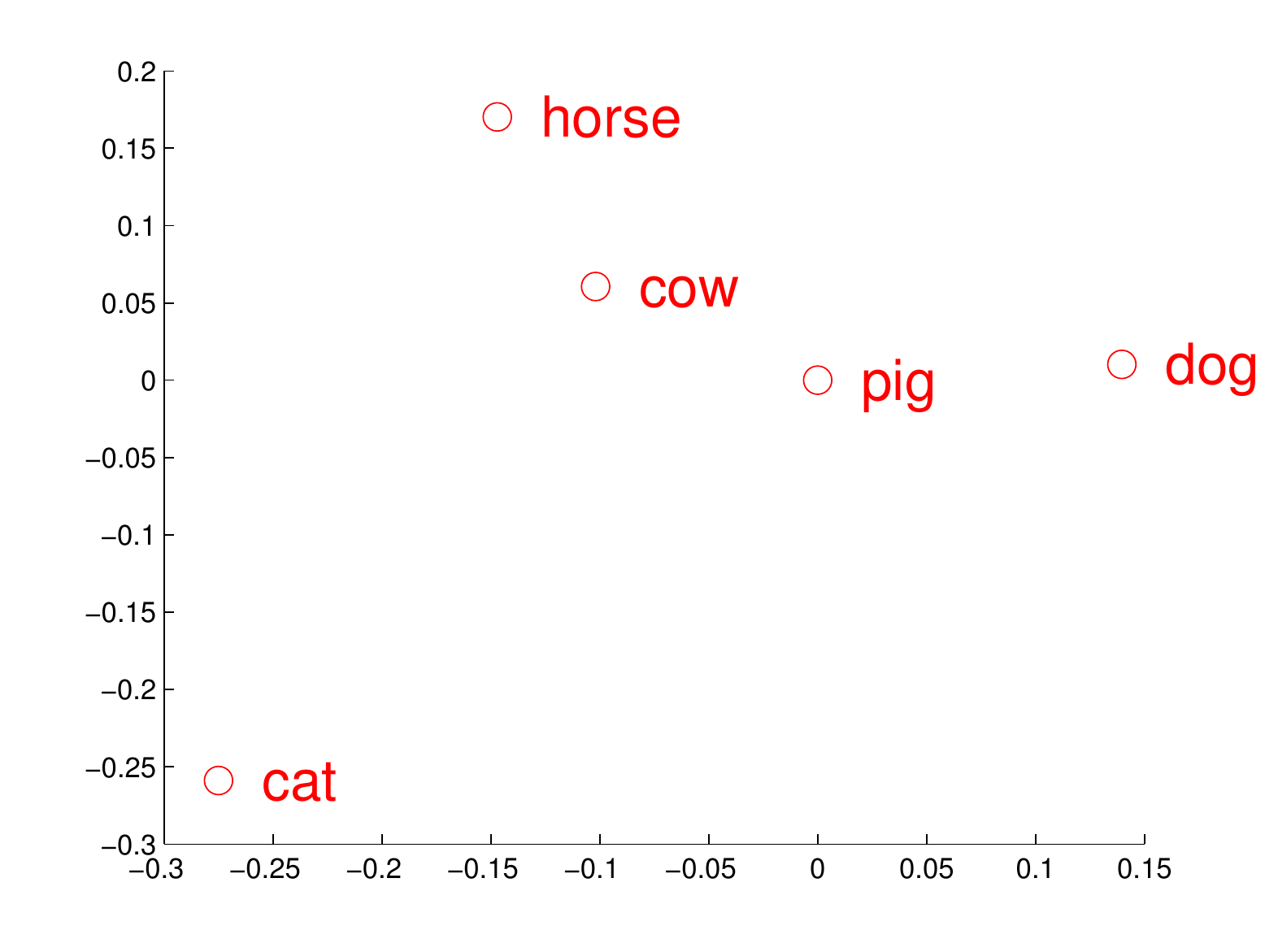}\hspace{.9cm}
\includegraphics[width=.9\columnwidth]{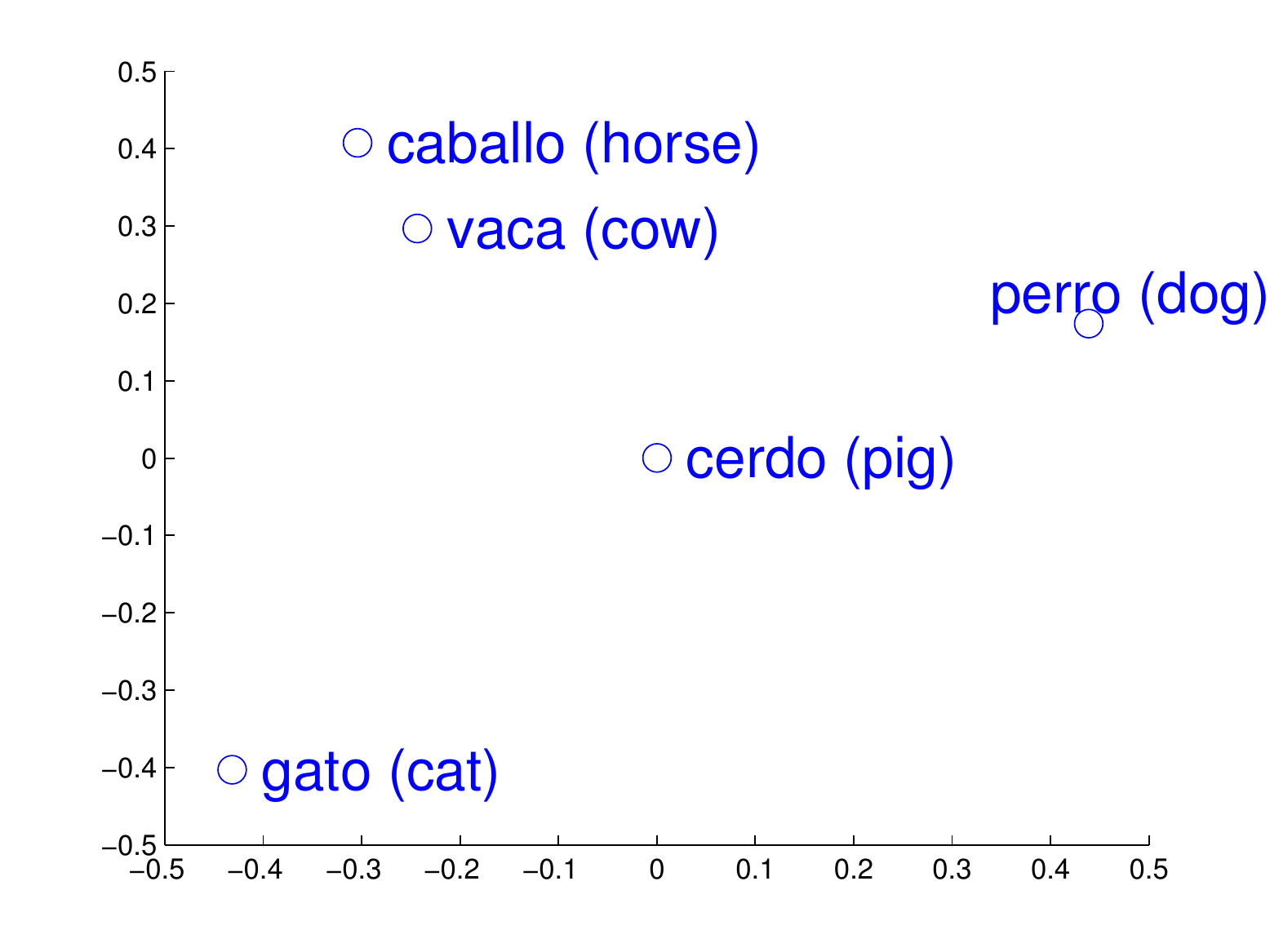}
\caption{Distributed word vector representations of numbers and
  animals in English (left) and Spanish (right). The five vectors in
  each language were projected down to two dimensions using PCA, and
  then manually rotated to accentuate their similarity. It can be
  seen that these concepts have similar geometric arrangements in both
  spaces, suggesting that it is possible to learn an accurate linear
  mapping from one space to another. This is the key idea behind our
  method of translation.\label{fig:linear-translation} }
\end{figure*}

Figure~\ref{fig:linear-translation} gives simple visualization to
illustrate why and how our method works. In
Figure~\ref{fig:linear-translation}, we visualize the vectors for
numbers and animals in English and Spanish, and it can be easily seen
that these concepts have similar geometric arrangements.  The reason
is that as all common languages share concepts that are grounded in
the real world (such as that cat is an animal smaller than a dog),
there is often a strong similarity between the vector spaces. The
similarity of geometric arrangments in vector spaces is the key reason
why our method works well.

Our proposed approach is complementary to the existing methods that use
similarity of word morphology between related languages
or exact context matches to infer the possible
translations~\cite{koehn2002learning,haghighi2008learning,koehn2000estimating}. Although
we found that morphological features (e.g., edit distance between word
spellings) can improve performance for related languages
(such as English to Spanish), our method is useful for
translation between languages that are substantially different (such as
English to Czech or English to Chinese).

Another advantage of our method is that it provides a translation
score for every word pair, which can be used in multiple ways. For
example, we can augment the existing phrase tables with more candidate
translations, or filter out errors from the translation tables and
known dictionaries.

\begin{figure*}[hbt]
\centering
\includegraphics[width=0.8\textwidth]{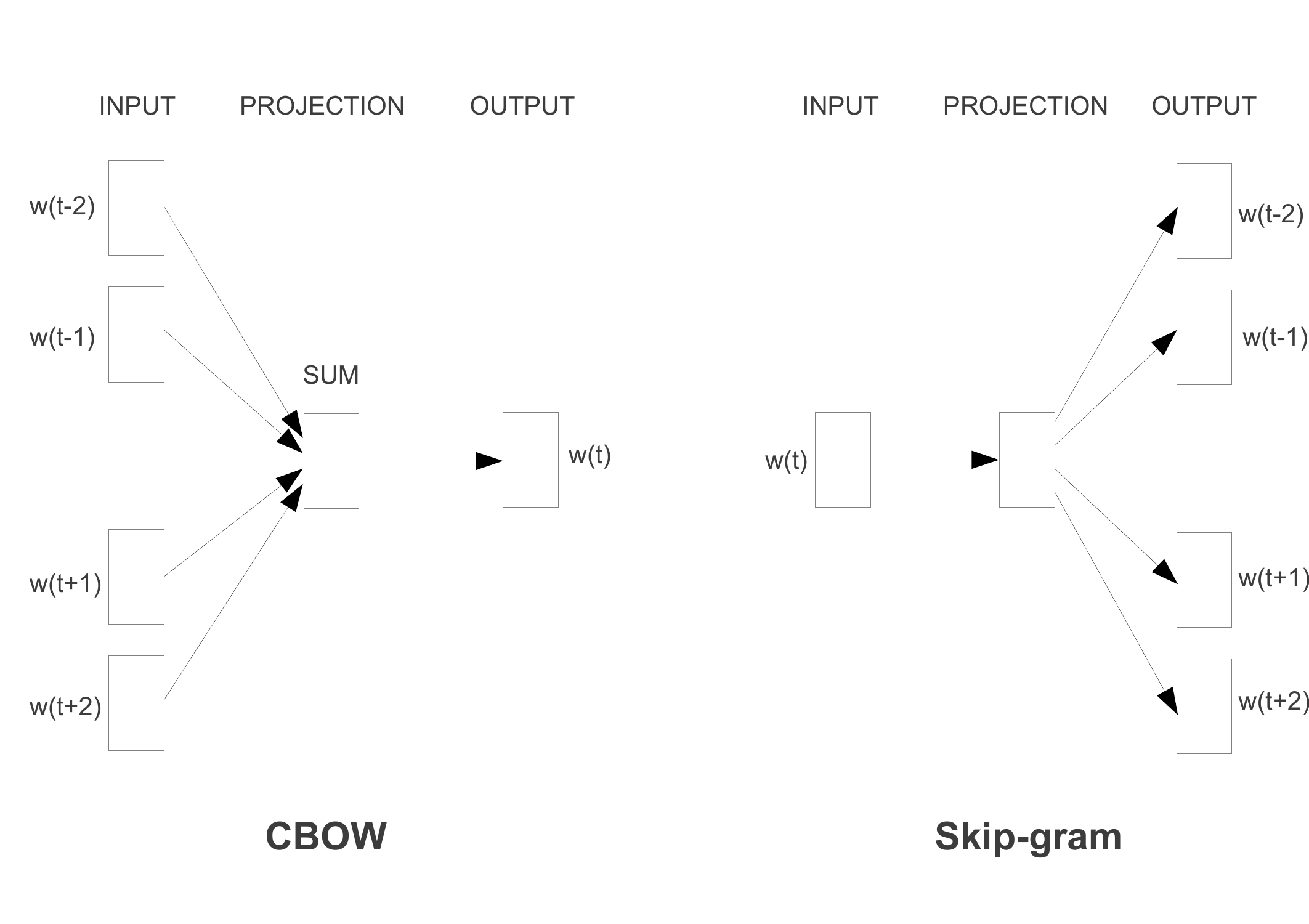}

\caption{Graphical representation of the CBOW model and Skip-gram
  model. In the CBOW model, the distributed representations of context
  (or surrounding words) are combined to predict the word in the
  middle. In the Skip-gram model, the distributed representation of
  the input word is used to predict the context. \label{fig:skipgram}}
\end{figure*}

\section{The Skip-gram and Continuous Bag-of-Words Models}

Distributed representations for words were proposed
in~\cite{rumelhart1986learning} and have become extremely successful.
The main advantage is that the representations of similar
words are close in the vector space, which makes generalization
to novel patterns easier and model estimation more robust.
Successful follow-up work includes applications to statistical language
modeling~\cite{elman,bengio,mikolov2012},
and to various other NLP tasks such as word representation learning, named entity
recognition, disambiguation, parsing, and tagging~\cite{collobert2008unified,turian2010word,socher2011parsing,socher,collobert2011natural,huang2012improving,mikolov}.

It was recently shown that the distributed representations of words capture surprisingly
many linguistic regularities, and that there are many types of similarities among words
that can be expressed as linear translations~\cite{mikolov2013naacl}. For example,
vector operations ``king'' - ``man'' + ``woman'' results in a vector that is close
to ``queen''.

Two particular models for learning word representations that can be
efficiently trained on large amounts of text data are Skip-gram and
CBOW models introduced in~\cite{mikolov}. In the CBOW model, the
training objective of the CBOW model is to combine the representations
of surrounding words to predict the word in the middle. The model
architectures of these two methods are shown in
Figure~\ref{fig:skipgram}. Similarly, in the Skip-gram model, the
training objective is to learn word vector representations that are
good at predicting its context in the same sentence~\cite{mikolov}. It
is unlike traditional neural network based language
models~\cite{bengio,mnih2008scalable,mikolov2010recurrent}, where the
objective is to predict the next word given the context of several
preceding words. Due to their low computational complexity, the
Skip-gram and CBOW models can be trained on a large corpus in a short
time (billions of words in hours). In practice, Skip-gram gives better
word representations when the monolingual data is small. CBOW however
is faster and more suitable for larger datasets~\cite{mikolov}. They
also tend to learn very similar representations for languages. Due to
their similarity in terms of model architecture, the rest of the
section will focus on describing the Skip-gram model.

More formally, given a
sequence of training words $w_1, w_2, w_3,\ldots,w_T$, the objective
of the Skip-gram model is to maximize the average log probability
\begin{equation}
\frac1T\sum_{t=1}^T\bigg[\sum_{j=-k}^k \log p(w_{t+j}|w_t)\bigg]
\end{equation}
where $k$ is the size of the training window (which can be a function
of the center word $w_t$). The inner summation goes from $-k$ to $k$
to compute the log probability of correctly predicting the word
$w_{t+j}$ given the word in the middle $w_t$. The outer summation goes
over all words in the training corpus.

In the Skip-gram model, every word $w$ is associated with two
learnable parameter vectors, $u_w$ and $v_w$. They are the ``input''
and ``output'' vectors of the $w$ respectively. The probability of
correctly predicting the word $w_i$ given the word $w_j$ is defined as
\begin{equation}
p(w_i|w_j) = \frac{\exp\left({u_{w_i}}^\top v_{w_j}\right)}
{\sum_{l=1}^{V}\exp\left({u_l}^\top v_{w_j}\right)}
\end{equation}
where $V$ is the number of words in the vocabulary.

This formulation is expensive because the cost of computing $\nabla
\log p(w_i|w_j)$ is proportional to the number of words in the
vocabulary $V$ (which can be easily in order of millions).  An efficient
alternative to the full softmax is the hierarchical softmax~\cite{hsoft_first},
which greatly reduces the complexity of computing $\log p(w_i|w_j)$ (about
logarithmically with respect to the vocabulary size).

The Skip-gram and CBOW models are typically trained using stochastic
gradient descent. The gradient is computed using backpropagation
rule~\cite{rumelhart1986learning}.

When trained on a large dataset, these models
capture substantial amount of semantic information. As mentioned
before, closely related words have similar vector representations, e.g.,
\emph{school} and \emph{university}, \emph{lake} and \emph{river}. This
is because \emph{school} and \emph{university} appear in
similar contexts, so that during training the vector representations of
these words are pushed to be close to each other.

More interestingly, the vectors capture relationships between concepts
via linear operations. For example, \emph{vector(France)} -
\emph{vector(Paris)} is similar to \emph{vector(Italy)} -
\emph{vector(Rome)}.

\section{Linear Relationships Between Languages}

As we visualized the word vectors using PCA, we noticed that the vector
representations of similar words in different languages were related
by a linear transformation. For instance,
Figure~\ref{fig:linear-translation} shows that the word vectors for
English numbers \emph{one} to \emph{five} and the corresponding
Spanish words \emph{uno} to \emph{cinco} have similar geometric
arrangements. The relationship between vector
spaces that represent these two languages can thus possibly be
captured by linear mapping (namely, a rotation and
scaling).

Thus, if we know the translation of \emph{one} and \emph{four} from
English to Spanish, we can learn the transformation matrix
that can help us to translate even the other numbers to Spanish.

\section{Translation Matrix}

Suppose we are given a set of word pairs and their associated vector
representations $\{x_{i}, z_{i}\}^{n}_{i=1}$, where $x_{i} \in
\mathbb{R}^{d_1}$ is the distributed representation of word $i$ in the
source language, and $z_{i} \in \mathbb{R}^{d_2}$ is the vector
representation of its translation.

It is our goal to find a transformation matrix $W$ such that $Wx_{i}$
approximates $z_{i}$. In practice, $W$ can be learned by the following
optimization problem
\begin{align}
\min_W \sum_{i=1}^n \| Wx_{i} - z_{i} \|^2
\end{align}
which we solve with stochastic gradient descent.

At the prediction time, for any given new word and its continuous vector
representation $x$, we can map it to the other language space by computing
$z = Wx$. Then we find the word whose representation is closest to $z$
in the target language space, using cosine similarity as the distance metric.

Despite its simplicity, this linear transformation method worked well
in our experiments, better than nearest neighbor and as well as neural network
classifiers. The following experiments will demonstrate its effectiveness.

\section{Experiments on WMT11 Datasets}

In this section, we describe the results of our translation
method on the publicly available WMT11 datasets. We also describe
two baseline techniques: one based on the edit distance between
words, and the other based on similarity of word co-occurrences that
uses word counts. The next section presents results on a larger dataset,
with size up to 25 billion words.

In the above section, we described two methods, Skip-gram and CBOW,
which have similar architectures and perform similarly. In terms of
speed, CBOW is usually faster and for that reason, we used it in
the following experiments.\footnote{It should be noted that the following
experiments deal mainly with frequent words. The
Skip-gram, although slower to train than CBOW, is preferable architecture
if one is interested in high quality represenations for the infrequent words.}

\subsection{Setup Description}

The datasets in our experiments are WMT11 text data from {\tt
www.statmt.org}
website.\footnote{http://www.statmt.org/wmt11/training-monolingual.tgz}
Using these corpora, we built monolingual data sets for English,
Spanish and Czech languages. We performed these steps:

\begin{itemize}
\item Tokenization of text using scripts from {\tt www.statmt.org}
\item Duplicate sentences were removed
\item Numeric values were rewritten as a single token
\item Special characters were removed (such as !?,:<)
\end{itemize}
Additionally, we formed short phrases of words using a technique described
in~\cite{phrases0}. The idea is that words that co-occur more
frequently than expected by their unigram probability are likely an
atomic unit. This allows us to represent short phrases such as ``ice
cream'' with single tokens, without blowing up the vocabulary size
as it would happen if we would consider all bigrams as phrases.

Importantly, as we want to test if our work can provide non-obvious translations of
words, we discarded named entities by removing the words
containing uppercase letters from our monolingual data. The named
entities can either be kept unchanged, or translated using simpler
techniques, for example using the edit
distance~\cite{koehn2002learning}. The statistics for the obtained
corpora are reported in Table~\ref{tab:stats}.

\begin{table}[thb]
\centering
\caption{The sizes of the monolingual training datasets from WMT11.
  The vocabularies consist of the words that occurred at least five
  times in the corpus.\vspace{3mm}}
\label{tab:stats}
\begin{tabular}{|l||l|l|}
\hline
\bf{Language} & \bf{Training tokens} & \bf{Vocabulary size} \\
\hline\hline
English            & 575M       & 127K \\
\hline
Spanish            &  84M       & 107K \\
\hline
Czech              & 155M       & 505K \\
\hline
\end{tabular}
\end{table}

\begin{table*}[bt]
\centering
\caption{Accuracy of the word translation methods using the WMT11 datasets. The Edit Distance uses morphological structure of words to find the translation. The Word Co-occurrence technique based on counts
uses similarity of contexts in which words appear, which is related to our proposed technique that uses continuous representations of words and a Translation Matrix between two languages.\vspace{3mm}}
\label{tab:wmt11results}
\begin{tabular}{|l||c|c||c|c||c|c||c|c||c|}
\hline
\bf{Translation}   & \multicolumn{2}{c||}{\bf Edit Distance} & \multicolumn{2}{c||}{\bf Word Co-occurrence}& \multicolumn{2}{c||}{\bf Translation Matrix} & \multicolumn{2}{c||}{\bf ED + TM}  & \bf{Coverage}  \\
                   & P@1        &  P@5                       & P@1    & P@5                                &           P@1       &      P@5               &   P@1        & P@5                 &                \\
\hline\hline
En $\to$ Sp        & 13\%       &  24\%                      & 19\%   & 30\%                               &           33\%      &      51\%              &   43\%       & 60\%                &        92.9\%  \\
\hline
Sp $\to$ En        & 18\%       &  27\%                      & 20\%   & 30\%                               &           35\%      &      52\%              &   44\%       & 62\%                &        92.9\%  \\
\hline
En $\to$ Cz        & 5\%        &  9\%                       & 9\%    & 17\%                               &           27\%      &      47\%              &   29\%       & 50\%                &        90.5\%  \\
\hline
Cz $\to$ En        & 7\%        &  11\%                      & 11\%   & 20\%                               &           23\%      &      42\%              &   25\%       & 45\%                &        90.5\%  \\
\hline
\end{tabular}
\end{table*}

To obtain dictionaries between languages, we used the most frequent words
from the monolingual source datasets, and translated these words using
on-line Google Translate (GT).
As mentioned previously, we also used
short phrases as the dictionary entries. As not all words that GT
produces are in our vocabularies that we built from the monolingual
WMT11 data, we report the vocabulary coverage in each experiment.
For the calculation of translation precision,
we discarded word pairs that cannot be translated
due to missing vocabulary entries.

To measure the accuracy, we use
the most frequent 5K words from the source language and their
translations given GT as the training data for learning the Translation
Matrix. The subsequent 1K words in the source language and their
translations are used as a test set. Because our approach is very good
at generating many translation candidates, we report the top 5
accuracy in addition to the top 1 accuracy. It should be noted that
the top 1 accuracy is highly underestimated, as synonym translations are
counted as mistakes - we count only exact match as a successful translation.

\subsection{Baseline Techniques}

We used two simple baselines for the further experiments,
similar to those previously described in~\cite{haghighi2008learning}. The first
is using similarity of the morphological structure of words, and
is based on the edit distance between words in different languages.

The second baseline uses similarity of word co-occurrences, and is thus
more similar to our neural network based approach. We follow these steps:
\begin{itemize}
\item Form count-based word vectors with dimensionality equal to the size of the dictionary
\item Count occurrence of in-dictionary words within a short window (up to 10 words) for each test word in the source language, and each word in the target language
\item Using the dictionary, map the word count vectors from the source language to the target language
\item For each test word, search for the most similar vector in the target language
\end{itemize}
Additionally, the word count vectors are normalized in the following
procedure. First we remove the bias that is introduced by the
different size of the training sets in both languages, by dividing the
counts by the ratio of the data set sizes.  For example, if we have
ten times more data for the source language than for the target
language, we will divide the counts in the source language by
ten. Next, we apply the log function to the counts and normalize each
word count vector to have a unit length (L2 norm).

The weakness of this technique is the computational complexity during the translation - the size of the word count
vectors increases linearly with the size of the dictionary, which makes the translation expensive.
Moreover, this approach ignores all the words that are not in the known dictionary when forming the count vectors.

\subsection{Results with WMT11 Data}

In Table~\ref{tab:wmt11results}, we report the performance of several
approaches for translating single words and short phrases.  Because the Edit Distance
and our Translation Matrix approach are fundamentally different, we
can improve performance by using a weighted combination of similarity
scores given by both techniques. As can be seen in
Table~\ref{tab:wmt11results}, the Edit Distance worked well for
languages with related spellings (English and Spanish), and
was less useful for more distant language pairs, such as English and Czech.

To train the distributed Skip-gram model, we used the
hyper-parameters recommended in~\cite{mikolov}: the window size is 10
and the dimensionality of the word vectors is in the hundreds. We
observed that the word vectors trained on the source language should
be several times (around 2x--4x) larger than the word vectors trained
on the target language. For example, the best performance on English
to Spanish translation was obtained with 800-dimensional English word
vectors and 200-dimensional Spanish vectors. However, for the opposite
direction, the best accuracy was achieved with 800-dimensional Spanish
vectors and 300-dimensional English vectors.

\section{Large Scale Experiments}

\begin{figure}
\centerline{
\includegraphics[width=1.1\columnwidth]{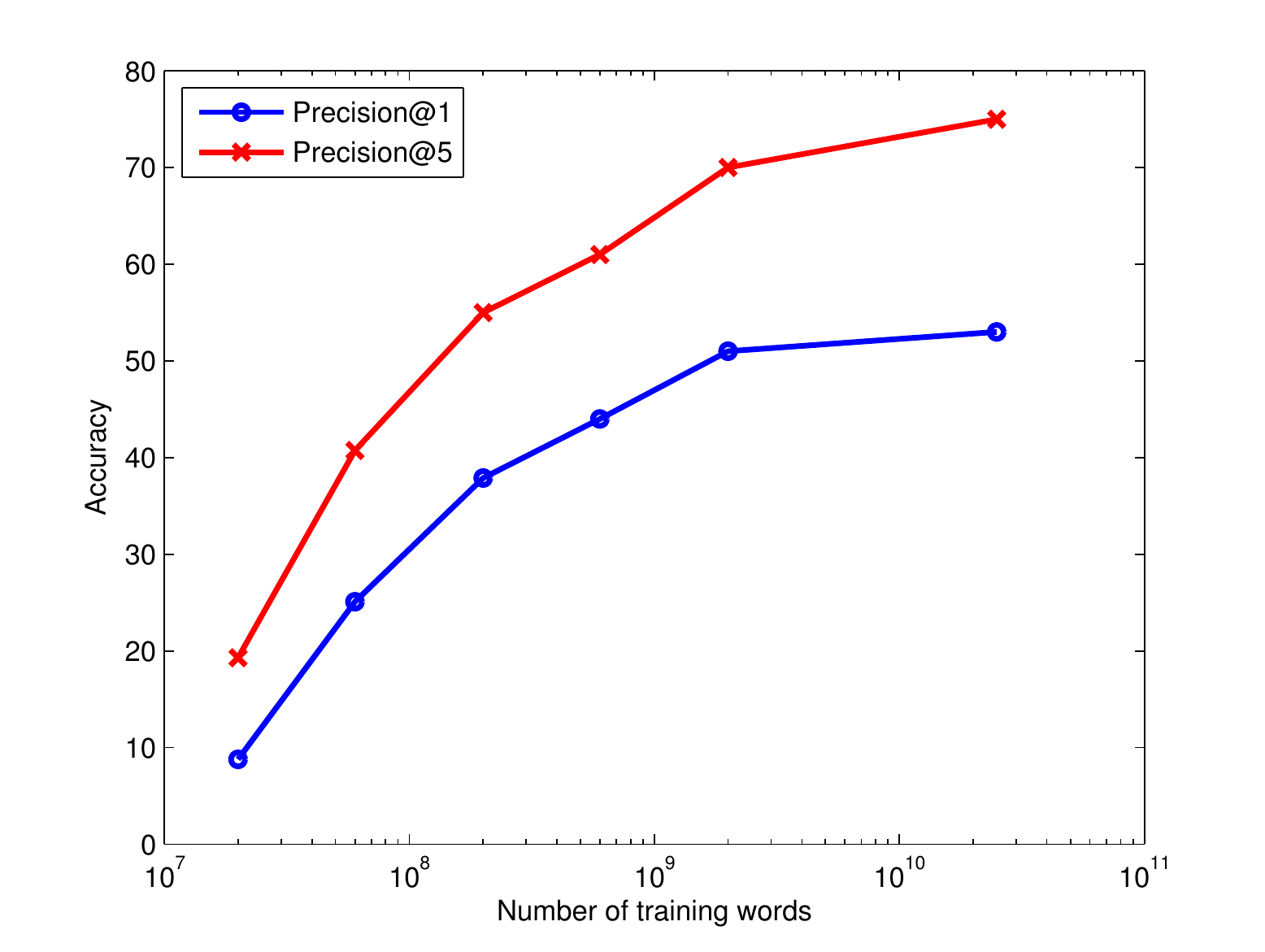}}
\caption{The Precision at 1 and 5 as the size of the
  monolingual training sets increase (EN$\to$ES).\label{fig:increase}}
\end{figure}

In this section, we scale up our techniques to larger datasets to show
how performance improves with more training data. For these experiments, we used large English
and Spanish corpora that have several
billion words (Google News datasets). We performed the same data cleaning and pre-processing
as for the WMT11 experiments. Figure~\ref{fig:increase} shows
how the performance improves as the amount of monolingual data
increases. Again, we used the
most frequent 5K words from the source language for constructing the
dictionary using Google Translate, and the next 1K words for test.

\begin{figure}
\includegraphics[width=1.05\columnwidth]{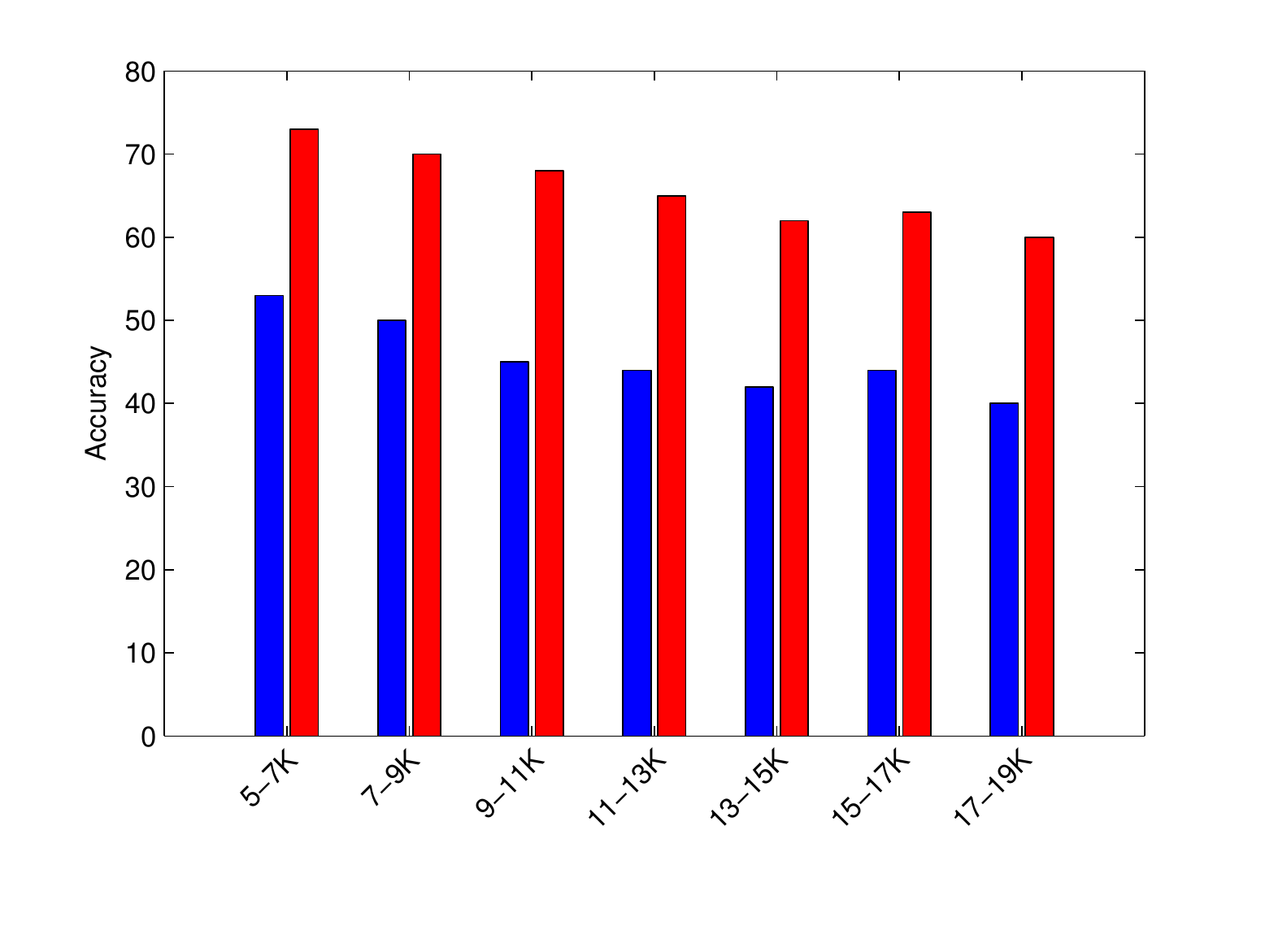}
\caption{Accuracies of translation as the word frequency
  decreases. Here, we measure the accuracy of the translation on
  disjoint sets of 2000 words sorted by frequency, starting from rank
  5K and continuing to 19K. In all cases, the linear transformation
  was trained on the 5K most frequent words and their translations.
  EN$\to$ES. \label{fig:freq}}
\end{figure}

Our approach can also be successfully used for translating infrequent
words: in Figure~\ref{fig:freq}, we show the translation accuracy on
the less frequent words. Although the translation accuracy decreases as
the test set becomes harder, for the words ranked 15K--19K in the
vocabulary, Precision@5 is still reasonably high, around 60\%. It is
surprising that we can sometimes translate even words that are
quite unrelated to those in the known dictionary. We will present examples
that demonstrate the translation quality in the Section~\ref{sec:ex}.

We also performed the same experiment where we translate the words at
ranks 15K--19K using the models trained on the small WMT11 datasets.
The performance loss in this case was greater--the Presicion@5 was
only 25\%. This means that the models have to be trained on large
monolingual datasets in order to accurately translate infrequent
words.

\subsection{Using Distances as Confidence Measure}

\begin{table}[b]
\centering
\caption{ Accuracy of our method using various confidence thresholds (EN$\to$ES, large corpora).
\vspace{3mm}}
\label{tab:methodsummary}
\begin{tabular}{|l|l|l|l|}
\hline
\bf{Threshold} & \bf{Coverage} & \bf{P@1} & \bf{P@5}\\
\hline\hline
0.0& 92.5\%   & 53\% & 75\%\\
\hline
0.5& 78.4\%   & 59\% & 82\%\\
\hline
0.6& 54.0\%   & 71\% & 90\%\\
\hline
0.7& 17.0\%   & 78\% & 91\%\\
\hline
\end{tabular}
\end{table}

\begin{table}[b]
\centering
\caption{ Accuracy of the combination of our method with the
  edit distance for various confidence thresholds. The confidence
  scores differ from the previous table since they include the edit
  distance (EN$\to$ES, large corpora).
\vspace{3mm}} 
\label{tab:methodsummary2}
\begin{tabular}{|l|l|l|l|}
\hline
\bf{Threshold} & \bf{Coverage} & \bf{P@1} & \bf{P@5}\\
\hline\hline
0.0& 92.5\%   & 58\% & 77\%\\
\hline
0.4& 77.6\%   & 66\% & 84\%\\
\hline
0.5& 55.0\%   & 75\% & 91\%\\
\hline
0.6& 25.3\%   & 85\% & 93\%\\
\hline
\end{tabular}
\end{table}

Sometimes it is useful to have higher accuracy at the expense of
coverage. Here we show that the distance between the computed vector
and the closest word vector can be used as a confidence measure.  If
we apply the Translation Matrix to a word vector in English and obtain
a vector in the Spanish word space that is not close to vector of
any Spanish word, we can assume that the translation is likely to be inaccurate.

More formally, we define the confidence score as $\max_{i\in V}\cos(Wx,z_i$),
and if this value is smaller than a threshold, the translation is skipped.
In Table~\ref{tab:methodsummary} we show how this approach is related
to the translation accuracy. For example, we can translate approximately half of the
words from the test set (on EN$\to$ES with frequency ranks 5K--6K)
with a very high Precision@5 metric (around 90\%).  By adding the edit
distance to our scores, we can further
improve the accuracy, especially for Precision@1 as is shown in
Table~\ref{tab:methodsummary2}.

These observations can be crucial in the future work, as we can see that
high-quality translations are possible for some subset of the vocabulary.
The idea can be applied also in the opposite way: instead of searching
for very good translations of missing entries in the dictionaries,
we can detect a subset of the existing dictionary that is likely to be
ambiguous or incorrect.

\section{Examples}
\label{sec:ex}
This section provides various examples of our translation method.

\subsection{Spanish to English Example Translations}

To better understand the behavior of our translation system, we show a
number of example translations from Spanish to English in
Table~\ref{tab:ex1}.  Interestingly, many of the mistakes are somewhat
meaningful and are semantically related to the correct translation.
These examples were produced by the translation matrix alone,
without using the edit distance similarity.

\begin{table}[thb]
\centering
\caption{Examples of translations of out-of-dictionary words from
  Spanish to English. The three most likely translations are
  shown. The examples were chosen at random from words at ranks
  5K--6K. The word representations were trained on the large corpora. \vspace{3mm}}
\label{tab:ex1}
\begin{tabular}{|l|l|l|}
\hline
\small
\bf{Spanish word} & \bf{Computed English} & \bf{Dictionary}\\
                  & \bf{Translations}      & \bf{Entry}     \\
\hline\hline
emociones         & emotions               & emotions       \\
                  & emotion                &                \\
                  & feelings               &                \\
\hline
protegida         & wetland                & protected      \\
                  & undevelopable          &                \\
                  & protected              &                \\
\hline
imperio           & dictatorship           & empire         \\
                  & imperialism            &                \\
                  & tyranny                &                \\
\hline
determinante      & crucial                & determinant    \\
                  & key                    &                \\
                  & important              &                \\
\hline
preparada         & prepared               & prepared       \\
                  & ready                  &                \\
                  & prepare                &                \\
\hline
millas            & kilometers             & miles          \\
                  & kilometres             &                \\
                  & miles                  &                \\
\hline
hablamos          & talking                & talk           \\
                  & talked                 &                \\
                  & talk                   &                \\
\hline
destacaron        & highlighted            & highlighted    \\
                  & emphasized             &                \\
                  & emphasised             &                \\
\hline
\end{tabular}
\end{table}

\begin{table}[tbh]
\centering
\caption{Examples of translations from English to Spanish with high confidence. The models were trained on the
large corpora. \vspace{3mm}}
\label{tab:ex2}
\begin{tabular}{|l|l|l|}
\hline
\small
\bf{English word} & \bf{Computed Spanish} & \bf{Dictionary}\\
                  & \bf{Translation}       & \bf{Entry}     \\
\hline\hline
pets              & mascotas               & mascotas       \\
mines             & minas                  & minas          \\
unacceptable      & inaceptable            & inaceptable    \\
prayers           & oraciones              & rezo           \\
shortstop         & shortstop              & campocorto     \\
interaction       & interacción            & interacción    \\
ultra             & ultra                  & muy            \\
beneficial        & beneficioso            & beneficioso    \\
beds              & camas                  & camas          \\
connectivity      & conectividad           & conectividad   \\
transform         & transformar            & transformar    \\
motivation        & motivaci\'{o}n         & motivaci\'{o}n \\
\hline
\end{tabular}
\end{table}

\begin{table}[h!]
\centering
\caption{Examples of translations where the dictionary entry and the
  output from our system strongly disagree. These examples were chosen
  manually to demonstrate that it might be possible to automatically
  find incorrect or ambiguous dictionary entries. The vectors were
  trained on the large corpora.}
\label{tab:ex3}
\vspace{3mm}
\begin{tabular}{|l|l|l|}
\hline
\small
\bf{English word} & \bf{Computed Czech}    & \bf{Dictionary} \\
                  & \bf{Translation}       & \bf{Entry}      \\
\hline\hline
said              & řekl                   & uvedený         \\
                  & {\it (said)}             & {\it (listed)}    \\
\hline
will              & může                   & vůle            \\
                  & {\it (can)}              & {\it (testament)} \\
\hline
did               & udělal                 & ano             \\
                  & {\it (did)}              & {\it (yes)}       \\
\hline
hit               & zasáhl                 & hit             \\
                  & {\it (hit)}              & {\it -}         \\
\hline
must              & musí                   & mošt            \\
                  & {\it (must)}             & {\it (cider)}     \\
\hline
current           & stávající              & proud           \\
                  & {\it (current)}          & {\it (stream)}    \\
\hline
shot              & vystřelil              & shot            \\
                  & {\it (shot)}             & {\it -}         \\
\hline
minutes           & minut                  & zápis           \\
                  & {\it (minutes)}          & {\it (enrollment)} \\
\hline
latest            & nejnovější             & poslední        \\
                  & {\it (newest)}           & {\it (last)}      \\
\hline
blacks            & černoši                & černá           \\
                  & {\it (black people)}     & {\it (black color)} \\
\hline
hub               & centrum                & hub             \\
                  & {\it (center)}           & {\it -}         \\
\hline
minus             & minus                  & bez             \\
                  & {\it (minus)}            & {\it (without)}   \\
\hline
retiring          & odejde                 & uzavřený        \\
                  & {\it (leave)}            & {\it (closed)}    \\
\hline
grown             & pěstuje                & dospělý         \\
                  & {\it (grow)}             & {\it (adult)}     \\
\hline
agents            & agenti                 & prostředky      \\
                  & {\it (agents)}           & {\it (resources)} \\
\hline
\end{tabular}
\end{table}

\subsection{High Confidence Translations}

In Table~\ref{tab:ex2}, we show the translations from English to Spanish with high confidence (score above 0.5).
We used both edit distance and translation matrix. As can be seen, the quality is very high, around 75\%
for Precision@1 as reported in Table~\ref{tab:methodsummary2}.

\subsection{Detection of Dictionary Errors}

A potential use of our system is the correction of dictionary errors.
To demonstrate this use case, we have trained the translation matrix
using 20K dictionary entries for translation between English and
Czech. Next, we computed the distance between the translation given
our system and the existing dictionary entry. Thus, we evaluate the
translation confidences on words from the training set.

In Table~\ref{tab:ex3}, we list examples where the distance between the
original dictionary entry and the output of the system was large. We
chose the examples manually, so this demonstration is highly
subjective. Informally, the entries in the existing dictionaries were
about the same or more accurate than our system in about 85\% of the cases, and
in the remaining 15\% our system provided better translation. Possible
future extension of our approach would be to train
the translation matrix on all dictionary entries except the one
for which we calculate the score.

\subsection{Translation between distant language pairs: English and Vietnamese}


The previous experiments involved languages that have a good
one-to-one correspondence between words. To show that our technique is
not limited by this assumption, we performed experiments on
Vietnamese, where the concept of a word is different than in English.

For training the monolingual Skip-gram model of Vietnamese,
we used large amount of
Google News data with billions of words. We performed the same data cleaning
steps as for the previous languages, and additionally automatically
extracted a large number of phrases using the technique described
in~\cite{phrases0}. This way, we obtained about 1.3B training
Vietnamese phrases that are related to English words and short phrases.
In Table~\ref{tab:vn}, we summarize the achieved results.

\begin{table}[thb]
\centering
\caption{The accuracy of our translation method between
  English and Vietnamese. The edit distance technique did not provide
  significant improvements. Although the accuracy seems low for the
  EN$\to$VN direction, this is in part due to the large number of synonyms in the VN model.
\vspace{3mm}}
\label{tab:vn}
\begin{tabular}{|l|l|l|l|}
\hline
\bf{Threshold} & \bf{Coverage} & \bf{P@1} & \bf{P@5}\\
\hline\hline
En $\to$ Vn  &    87.8\%  &  10\%    &   30\%   \\
\hline
Vn $\to$ En  &    87.8\%  &  24\%    &   40\%   \\
\hline
\end{tabular}
\end{table}

\section{Conclusion}

In this paper, we demonstrated the potential of distributed
representations for machine translation. Using large amounts of
monolingual data and a small starting dictionary, we can successfully
learn meaningful translations for individual words and short
phrases. We demonstrated that this approach works well even for pairs
of languages that are not closely related, such as English and Czech,
and even English and Vietnamese.

In particular, our work can be used to enrich and improve
existing dictionaries and phrase tables, which would in turn lead to improvement
of the current state-of-the-art machine translation systems.
Application to low resource domains is another very interesting
topic for future research. Clearly, there is still much to be explored.

\bibliographystyle{naaclhlt2013}
\bibliography{translate}

\begin{thebibliography}{}

\bibitem[\protect\citename{Bengio \bgroup et al.\egroup }2003]{bengio}
Yoshua Bengio, Rejean Ducharme, Pascal Vincent, and Christian Jauvin.
\newblock 2003.
\newblock A neural probabilistic language model.
\newblock In {\em Journal of Machine Learning Research}, pages 1137--1155.

\bibitem[\protect\citename{Collobert and Weston}2008]{collobert2008unified}
Ronan Collobert and Jason Weston.
\newblock 2008.
\newblock A unified architecture for natural language processing: Deep neural
  networks with multitask learning.
\newblock In {\em Proceedings of the 25th international conference on Machine
  learning}, pages 160--167. ACM.

\bibitem[\protect\citename{Collobert \bgroup et al.\egroup
  }2011]{collobert2011natural}
Ronan Collobert, Jason Weston, L{\'e}on Bottou, Michael Karlen, Koray
  Kavukcuoglu, and Pavel Kuksa.
\newblock 2011.
\newblock Natural language processing (almost) from scratch.
\newblock {\em The Journal of Machine Learning Research}, 12:2493--2537.

\bibitem[\protect\citename{Elman}1990]{elman}
Jeff Elman.
\newblock 1990.
\newblock Finding structure in time.
\newblock In {\em Cognitive Science}, pages 179--211.

\bibitem[\protect\citename{Haghighi \bgroup et al.\egroup
  }2008]{haghighi2008learning}
Aria Haghighi, Percy Liang, Taylor Berg-Kirkpatrick, and Dan Klein.
\newblock 2008.
\newblock Learning bilingual lexicons from monolingual corpora.
\newblock In {\em ACL}, volume 2008, pages 771--779.

\bibitem[\protect\citename{Huang \bgroup et al.\egroup
  }2012]{huang2012improving}
Eric Huang, Richard Socher, Christopher Manning, and Andrew~Y Ng.
\newblock 2012.
\newblock Improving word representations via global context and multiple word
  prototypes.
\newblock In {\em Proceedings of the 50th Annual Meeting of the Association for
  Computational Linguistics: Long Papers-Volume 1}, pages 873--882. Association
  for Computational Linguistics.

\bibitem[\protect\citename{Koehn and Knight}2000]{koehn2000estimating}
Philipp Koehn and Kevin Knight.
\newblock 2000.
\newblock Estimating word translation probabilities from unrelated monolingual
  corpora using the em algorithm.
\newblock In {\em AAAI/IAAI}, pages 711--715.

\bibitem[\protect\citename{Koehn and Knight}2002]{koehn2002learning}
Philipp Koehn and Kevin Knight.
\newblock 2002.
\newblock Learning a translation lexicon from monolingual corpora.
\newblock In {\em Proceedings of the ACL-02 workshop on Unsupervised lexical
  acquisition-Volume 9}, pages 9--16. Association for Computational
  Linguistics.

\bibitem[\protect\citename{Mikolov \bgroup et al.\egroup
  }2010]{mikolov2010recurrent}
Tomas Mikolov, Martin Karafi{\'a}t, Lukas Burget, Jan Cernock{\`y}, and Sanjeev
  Khudanpur.
\newblock 2010.
\newblock Recurrent neural network based language model.
\newblock In {\em INTERSPEECH}, pages 1045--1048.

\bibitem[\protect\citename{Mikolov \bgroup et al.\egroup }2013a]{mikolov}
Tomas Mikolov, Kai Chen, Greg Corrado, and Jeffrey Dean.
\newblock 2013a.
\newblock Efficient estimation of word representations in vector space.
\newblock {\em arXiv preprint arXiv:1301.3781}.

\bibitem[\protect\citename{Mikolov \bgroup et al.\egroup }2013b]{phrases0}
Tomas Mikolov, Ilya Sutskever, Kai Chen, Greg Corrado, and Jeffrey Dean.
\newblock 2013b.
\newblock Distributed representations of phrases and their compositionality.
\newblock In {\em NIPS}.

\bibitem[\protect\citename{Mikolov \bgroup et al.\egroup
  }2013c]{mikolov2013naacl}
Tomas Mikolov, Scott Wen-tau Yih, and Geoffrey Zweig.
\newblock 2013c.
\newblock Linguistic regularities in continuous space word representations.
\newblock In {\em NAACL HLT}.

\bibitem[\protect\citename{Mikolov}2012]{mikolov2012}
Tomas Mikolov.
\newblock 2012.
\newblock {\em Statistical Language Models based on Neural Networks}.
\newblock {Ph.D.} thesis, Brno University of Technology.

\bibitem[\protect\citename{Mnih and Hinton}2008]{mnih2008scalable}
Andriy Mnih and Geoffrey~E Hinton.
\newblock 2008.
\newblock A scalable hierarchical distributed language model.
\newblock In {\em Advances in neural information processing systems}, pages
  1081--1088.

\bibitem[\protect\citename{Morin and Bengio}2005]{hsoft_first}
Frederic Morin and Yoshua Bengio.
\newblock 2005.
\newblock Hierarchical probabilistic neural network language model.
\newblock In {\em Proceedings of the international workshop on artificial
  intelligence and statistics}, pages 246--252.

\bibitem[\protect\citename{Rumelhart \bgroup et al.\egroup
  }1986]{rumelhart1986learning}
David~E Rumelhart, Geoffrey~E Hinton, and Ronald~J Williams.
\newblock 1986.
\newblock Learning representations by back-propagating errors.
\newblock {\em Nature}, 323(6088):533--536.

\bibitem[\protect\citename{Socher \bgroup et al.\egroup
  }2011]{socher2011parsing}
Richard Socher, Cliff~C Lin, Andrew Ng, and Chris Manning.
\newblock 2011.
\newblock Parsing natural scenes and natural language with recursive neural
  networks.
\newblock In {\em Proceedings of the 28th International Conference on Machine
  Learning (ICML-11)}, pages 129--136.

\bibitem[\protect\citename{Socher \bgroup et al.\egroup }2013]{socher}
Richard Socher, John Bauer, Christopher~D. Manning, and Andrew~Y. Ng.
\newblock 2013.
\newblock Parsing with compositional vector grammars.
\newblock In {\em {ACL}}.

\bibitem[\protect\citename{Turian \bgroup et al.\egroup }2010]{turian2010word}
Joseph Turian, Lev Ratinov, and Yoshua Bengio.
\newblock 2010.
\newblock Word representations: a simple and general method for semi-supervised
  learning.
\newblock In {\em Proceedings of the 48th Annual Meeting of the Association for
  Computational Linguistics}, pages 384--394. Association for Computational
  Linguistics.

\end{thebibliography}
\end{document}